\documentclass[11pt,letterpaper]{article}
\usepackage{emnlp2016}
\usepackage{times}
\usepackage{latexsym}
\usepackage{booktabs}

\usepackage{amsfonts} 

\usepackage[pdftex]{graphicx}
\graphicspath{{figures/}}
\DeclareGraphicsExtensions{.pdf,.jpeg,.png}

\emnlpfinalcopy

\def\emnlppaperid{***}

\title{Modelling Radiological Language with Bidirectional Long Short-Term Memory Networks}

\author{Savelie Cornegruta, Robert Bakewell, Samuel Withey \and Giovanni Montana\\
Department of Biomedical Engineering, King's College London, UK\\
  {\tt giovanni.montana@kcl.ac.uk}}

\date{}

\usepackage[small,compact]{titlesec} 

\begin{document}

\maketitle

\begin{abstract}
Motivated by the need to automate medical information extraction from free-text radiological reports, we present a bi-directional long short-term memory (BiLSTM) neural network architecture for modelling radiological language. The model has been used to address two NLP tasks: medical named-entity recognition (NER) and negation detection. We investigate whether learning several types of word embeddings improves BiLSTM's performance on those tasks. Using a large dataset of chest x-ray reports, we compare the proposed model to a baseline dictionary-based NER system and a negation detection system that leverages the hand-crafted rules of the NegEx algorithm and the grammatical relations obtained from the Stanford Dependency Parser. Compared to these more traditional rule-based systems, we argue that BiLSTM offers a strong alternative for both our tasks. 
\end{abstract}

\section{Introduction}


Radiological reports represent a large part of all Electronic Medical Records (EMRs) held by medical institutions. For instance, in England alone, upwards of 22 million plain radiographs were reported over the 12-month period from March 2015 \cite{nhs2016diagnostic}. A radiological report is a written document produced by a Radiologist, a physician that specialises in interpreting medical images. A report typically states any technical factors relevant to the acquired image as well as the presence or absence of radiological abnormalities. When an abnormality is noted, the Radiologist often gives further description, including anatomical location and the extent of the disease. 

Whilst Radiologists are taught to review radiographs in a systematic and comprehensive manner, their reporting style can vary quite dramatically \cite{reiner2006radiology} and the same findings can often be described in a multitude of different ways \cite{sobel1996information}. The radiological reports may contain broken grammar and misspellings, which are often the result of voice recognition software or the dictation-transcript method \cite{mcgurk2014effect}. Applying text mining techniques to these reports poses a number of challenges due to extensive variability in language, ambiguity and uncertainty, which are typical problems for natural language.


In this work we are motivated by the need to automatically extract standardised clinical information from digitised radiological reports. A system for the fully-automated extraction of this information could be used, for instance, to characterise the patient population and help health professionals improve day-to-day services. The extracted structured data could also be used to build management dashboards \cite{simpao2014review} summarising and presenting the most prevalent conditions. Another potential use is the automatic labelling of medical images, e.g. to support the development of computer-aided diagnosis software \cite{shin2015interleaved}.

In this paper we propose a recurrent neural network (RNN) architecture for modelling radiological language and investigate its potential advantages on two different tasks: medical named-entity recognition (NER) and negation detection. The model, a bi-directional long short-term memory (BiLSTM) network, does not use any hand-engineered features, but learns them using a relatively small amount of labelled data and a larger but unlabelled corpus of radiological reports. In addition, we explore the combined use of BiLSTM with other language models such as GloVe \cite{pennington2014glove} and a novel variant of GloVe, proposed here, that makes use of a medical ontology. The performance of the BiLSTM model is assessed comparatively to a rule-based system that has been optimised for the tasks at hand and builds upon well established techniques for medical NER and negation detection. In particular, for NER, the system uses a baseline dictionary-based text mining component relying on a curated dictionary of medical terms. As a baseline for the negation detection task, the system implements a hybrid component based on the NegEx algorithm \cite{chapman2013extending} in conjunction with grammatical relations obtained from the Stanford Dependency Parser \cite{chen2014fast}.

The article is organised as follows. In Section \ref{related_work} we provide a brief review of the existing body of work in NLP for medical information extraction and briefly discuss the use of artificial neural networks for NLP tasks. In Section \ref{data} we describe the datasets used for our experiments, and in Section \ref{methodology} we introduce the BiLSTM model. The results are presented in Section \ref{results} where we also compare BiLSTM against the rule-based baseline systems described in Section \ref{rule_based}.

\section{Related Work} \label{related_work}

\subsection{Medical NER}

A large proportion of NLP systems for medical text mining use dictionary-based methods for extracting medical concepts from clinical document \cite{friedman1995natural,Johnson:1997:EIF:2737436.2737540,aronson2001effective,savova2010mayo}. The dictionaries that contain the correspondence between a single- or multi-word phrase and a medical concept are usually built from medical ontologies such as the Unified Medical Language System (UMLS) \cite{umls} and Medical Subject Headings (MeSH) \cite{mesh}. These ontologies contain hundreds of thousands of medical concepts. There are also domain-specific ontologies such as RadLex \cite{langlotz2006radlex}, which has been developed for the Radiology domain, and currently contains over $68,000$ concepts.

Medical Language Extraction and Encoding System (MEDLEE) \cite{friedman1995natural} is one of the earliest automated systems originally developed for handling radiological reports, and later expanded to other medical domains. MEDLEE parses the given clinical documents by string matching: the words are matched to a pre-defined dictionary of medical terms or semantic groups (e.g. \textit{Central Finding}, \textit{Bodyloc Modifier}, \textit{Certainty Modifier} and \textit{Region Modifier}). Once the words have been associated with a semantic group, a Compositional Regularizer stage combines them according to a list of pre-defined mappings to form regularized multi-word phrases. The final stage looks up the regularized terms in a dictionary of medical concepts (e.g. \textit{enlarged heart} is mapped to the corresponding concept \textit{cardiomegaly}). A separate study evaluated MEDLEE on $150$ manually annotated radiology reports \cite{hripcsak2002use}; MEDLEE was assessed on its ability to detect $24$ clinical conditions achieving an average sensitivity and specificity of $0.81$ and $0.99$, respectively.



A more recent system for general medical information extraction is the Mayo Clinic's Text Analysis and Knowledge Extraction System (cTAKES) \cite{savova2010mayo}, which also implements an NLP pipeline. During an initial shallow parsing stage, cTAKES attempts to group words into multi-word expressions by identifying constituent parts of the sentence (e.g. noun, prepositional, and verb phrases). It then string matches the identified phrases to a concept in UMLS. A new set of semantic groups were also derived from the UMLS ontology \cite{ogren2007constructing}. The NER performance of the cTAKES was evaluated on the semantic groups, achieving an F1-score of $0.715$ for exact matches and $0.824$ for overlapping matches. 


In general, dictionary-based systems perform with high precision on the NER tasks but have a low recall, showing a lack of generalisation. Low recall is usually caused by the inability to identify multi-word phrases as concepts, unless exact matches can be found in the dictionary. In addition, such systems are not able to easily deal with disjoint entities. For instance, in the phrase {\it lungs are mildly hyperexpanded}, {\it hyperexpanded lungs} constitutes a clinical finding. In an attempt to deal with disjoint entities, rule-based systems such as MEDLEE, MetaMap \cite{aronson2001effective} and cTAKES, implement additional parsing stages to find grammatical relations between different words in a sentence, thus aiming to create disjoint multi-word phrases. However, state-of-the-art syntactic parsers are still likely to fail when parsing sentences with broken grammar, as often occurs in clinical documents.



In an attempt to improve upon dictionary-based information extraction systems, Hassanpour \shortcite{hassanpour2015information} recently used a first-order linear-chain Conditional Random Field (CRF) model \cite{lafferty2001conditional} in a medical NER task involving five semantic groups (anatomy, anatomy modifier, observation, observation modifier, and uncertainty). The features used for the CRF model included part-of-speech (POS) tags, word stems, word n-grams, word shape, and negations extracted using the NegEx algorithm. The model was trained and tested using $10$-fold cross validation on a corpus of $150$ multi-institutional Radiology reports and achieved a precision score of $0.87$, recall of $0.84$, and F1-score of $0.85$. 



\subsection{Medical negation detection}

NegEx, a popular negation detection algorithm, is usually applied to medical concepts after the entity recognition stage. This tool uses a curated list of phrases (e.g. \textit{no}, \textit{no sign of}, \textit{free of}), which are string matched to the medical text to detect a negation trigger, i.e. a word or phrase indicating the presence of a negated medical entity in the sentence. The target entities falling inside a window, starting at the negation trigger, are then classified as \textit{negated}. 
In light of its simplicity, speed and reasonable results, NegEx had been used as a component by many medical NLP systems \cite{wu2014negation}. It has been shown that that NegEx achieves an accuracy of $0.94$ as part of the cTAKES evaluation \cite{savova2010mayo}. However, the window approach that is used for classifying the negations may result in a large number of false positives, especially if there are multiple entities within the $6$-word window.

Aiming to reduce the number of false positives, recent efforts have integrated NegEx with machine learning models that can be trained on annotated datasets. For instance, Shivade \shortcite{shivade2015extending} introduced a kernel-based approach that uses features built using the type of negation trigger, features that are derived from the existence of conjunctions in the sentence, and features that weight the NegEx output against the bag-of-words in the dataset. The kernel based model outperformed the original NegEx algorithm by $2.7$ F1-score points when trained and tested on the NegEx dataset. At around the same time, Mehrabi \shortcite{mehrabi2015deepen} introduced DEEPEN, an algorithm that filters the NegEx output using the grammatical relations extracted using Stanford Dependency Parser. DEEPEN succeeded at reducing the number of false positives, although it showed a marginally lower F1-score when compared with NegEx on concepts from the \textit{Disorders} semantic group from the Mayo Clinic dataset \cite{ogren2007constructing}.



\subsection{Neural networks for NLP tasks}

In recent years, deep artificial neural networks have been found to yield consistently good results on various NLP tasks. The SENNA system \cite{collobert2011natural}, which used a convolutional neural network (CNN) architecture, came close to achieving state-of-the-art performance across the tasks of POS tagging, shallow parsing, NER, and semantic role labeling. More recently, recurrent neural networks (RNNs) have been shown to achieve very high performance, and often reach state-of-the-art results in various language modelling tasks \cite{mikolov2012context}. RNNs have also been shown to outperform more traditional machine learning models, such as Logistic Regression and CRF, at the slot filling task in spoken language understanding \cite{mesnil2013investigation}. In a NER task on the publicly available datasets in four languages, the bidirectional long short-term memory (LSTM) networks \cite{hochreiter1997long}, a variant of RNN, outperformed CNNs, CRFs and other models \cite{lample2016neural}. 



Neural networks have also been used to learn language models in an unsupervised learning setting. Some popular models include Skip-gram and continuous bag-of-words (CBOW) \cite{mikolov2013distributed}. These yield word representations, or embeddings, that are able to carry the syntactic and semantic information of a language. Collobert \shortcite{collobert2011natural} showed that integrating pre-trained word embeddings into a neural network can help the supervised learning process.

\section{A Radiology corpus} \label{data}

\subsection{Dataset} \label{dataset}

For this study, we produced an in-house radiology corpus consisting of $745,480$ historical chest X-ray (radiographs) reports provided by Guy's and St Thomas' Trust (GSTT). This Trust runs two hospitals within the National Health Service (NHS) in England, serving a large area in South London. The reports cover the period between January $2005$ and March $2016$, and were generated by $276$ different reporters including consultant Radiologists, trainee Radiologists and reporting Radiographers. Our repository consists of text written or dictated by the clinicians after radiograph analysis, and do not contain any referral information or patient-identifying data, such as names, addresses or dates of birth. However, many reports refer to the clinical history of the patient. The reports had a minimum of $1$ word and maximum of $311$ words, with an average of $25.3$ words and a standard deviation of $19.9$ words. On average there were $2.9$ sentences per report. After lemmatization, converting to lower case, and discounting words that occur less than $3$ times in the corpus, the resulting vocabulary contained $8,031$ words. 

\begin{figure}[!t]
\centering
\includegraphics[width=\columnwidth]{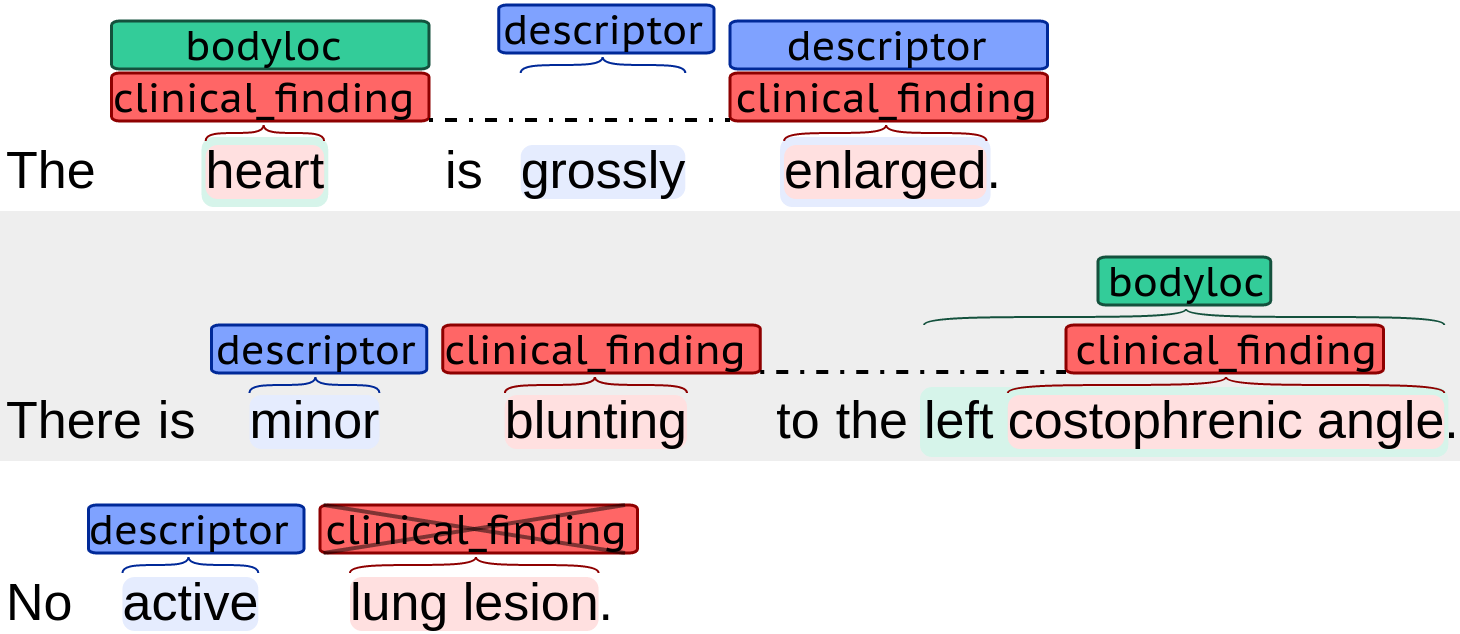}
\caption{Example of manual annotation of a radiology report performed using BRAT}
\label{bratexample}
\end{figure}

A sample of $2,000$ reports was randomly selected from the corpus for the purpose of creating a training and validation dataset for the NER and negation detection tasks, whilst the remaining of the reports were utilised for pre-training word embeddings. The reports selected for manual annotation were written for all types of patients (Inpatient: $1072$, A\&E Attender: $515$, Outpatient: $229$, GP Direct Access Patient: $165$, Ward Attender: $9$, Day Case Patient: $8$) by $144$ different clinicians.

We introduce a simple word-level annotation schema that includes four classes or semantic groups: {\it Clinical Finding}, {\it Body Location}, {\it Descriptor} and {\it Medical Device}: Clinical Finding encompasses any clinically-relevant radiological abnormality, Body Location refers to the anatomical area where the finding is present, and Descriptor includes all adjectives used to describe the other classes. The Medical Device class is used to label any medical apparatus seen on chest radiographs, such as pacemakers, intravascular lines, and nasogastric tubes. Our annotation schema allows for the same token to belong to several semantic groups. For example, as shown in Figure \ref{bratexample}, the word \textit{heart} was associated with both Clinical Finding and Body Location classes. We have also introduced a negation attribute to indicate the absence of any of these entities. 

\subsection{Gold standard}

Two clinicians (RB and SW) annotated the reports using BRAT \cite{stenetorp2012brat}, a collaborative tool for text annotation that was configured to use our own schema. The BRAT output was then transformed to the IOBES tagging schema. Here, we interpret I as a token in the middle of an entity; O as a token not part of the entity; B and E as the beginning and end of the entity, respectively; finally, S indicates a single-word entity. We work with the assumption that entities may be disjoint and tokens that are surrounded by disjoint entity may belong to a different semantic group. For example, according to the annotation performed by the clinicians, in the sentence \textit{Heart is slightly enlarged} the phrase \textit{heart enlarged} represents an entity that belongs to the semantic group \textit{Clinical Finding} and \textit{slightly} is a \textit{Descriptor}. The resulting breakdown of all entities by semantic group can be found in Table \ref{anno_summary}. 


\begin{table}[!t]
\small
\centering
\begin{tabular}{@{}ccc@{}}
\toprule
\textbf{Semantic Group} & \textbf{\# of entities} & \textbf{\# of tokens} \\ \midrule
Body Location           & 5686                        & 10113                     \\
Clinical Finding        & 5396                        & 8906                      \\
Descriptor              & 3458                        & 3845                      \\
Medical Device          & 1711                        & 3361                      \\ \midrule
Total                   & 16251                       & 26225                     \\ \midrule
Negated entities        & 1851                        & 2557                      \\ \bottomrule
\end{tabular}
\caption{Frequency distribution of entities by class in $2,000$ manually annotated reports}
\label{anno_summary}
\end{table}

\section{Methodology} \label{methodology}




In this Section we describe a model for NER that extracts five types of entities: the four semantic groups described in Section \ref{dataset}, as well as the negation, which is treated here as an additional class, analogously to the semantic groups.

\subsection{Bi-directional LSTM}

The RNN is a neural network architecture designed to model time series, but it can be applied to other types of sequential data \cite{rumelhart1988learning}. As the information passes through the network, it can persist indefinitely in its  memory. This facilitates the process of capturing sequential dependencies. The RNN makes a prediction after processing each element of the input sequence. Hence, the output sequence can be of the same length as the input sequence. The RNN architecture lends itself as a natural model for the proposed NER task, where the objective is to predict the IOBES tags for each of the input words.


The RNN is trained using the error backpropagation through time algorithm \cite{werbos1990backpropagation} and a variant of the gradient descent algorithm. However, training these models is notoriously challenging due to the problem of exploding and vanishing gradients, especially when trained with long input sequences \cite{bengio1994learning}. For the exploding gradient problem, numerical stability can be achieved by clipping the gradients \cite{graves2013generating}. The problem of vanishing gradients can be addressed by replacing the standard RNN cell with a long short-term memory (LSTM) cell, which allows for a constant error flow along the input sequence \cite{hochreiter1997long}. A more constant error also means that the network is able to learn better long-term dependencies over the input sequence. By combining the outputs of two RNNs that pass the information in opposing directions, it is possible to capture the context from both ends of the sequence. The resulting architecture is known as Bidirectional LSTM (BiLSTM) \cite{graves2005framewise}.

\begin{figure}[!t]
 \centering 
 \includegraphics[width=\columnwidth]{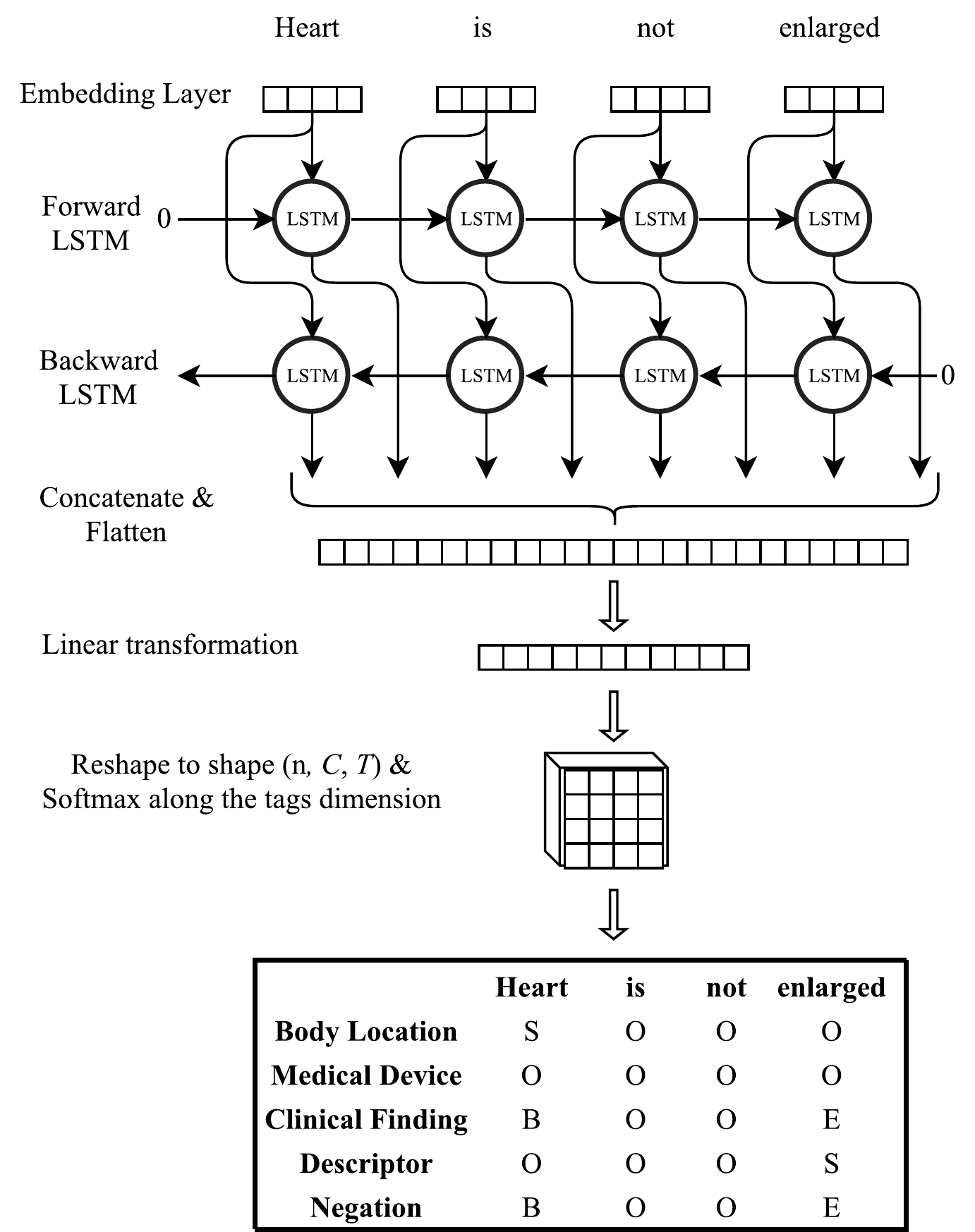}
 \caption{An illustration of the BiLSTM architecture for joint medical entity recognition and negation detection}
  \label{model_diagram}
\end{figure}

We start by defining a vocabulary ${V=\{v_1, v_2, ..., v_{8031}\}}$ that contains the words extracted from the corpus as described in Section \ref{dataset}. We assume that, in order to perform NER on the words in any given sentence, it is sufficient to consider only the information contained in that sentence. Therefore we pass the BiLSTM one sentence at a time. For each input sentence of $n$ words we define an $n$-dimensional vector $\mathbf{x}$ whose elements are the indices in $V$ corresponding to words appearing in the sentence, preserving the order. The input $\mathbf{x}$ is passed to an Embedding Layer that returns the sequence ${S=\{w_j | j = x_1, x_2, ..., x_n \}}$ where $w_j$ is the $j$th row of a dense matrix ${\mathbf{W} \in \mathbb{R}^{|V| \times d}}$, where ${d \in \mathbb{N}}$ is a hyperparameter. The vector $w_j$ represents a low-dimensional vector representation, or word embedding, whereas $\mathbf{W}$ is the corresponding embedding matrix. The sequence of word embeddings $S$ is then passed as input to two LSTM layers that process it in opposing directions (forwards and backwards), similar to the architecture introduced by Graves \shortcite{graves2005framewise}. Figure \ref{model_diagram} shows the LSTM layers in their "unrolled" form as they read the input. Each LSTM layer contains $k$ LSTM memory cells which are based on the implementation by Graves \shortcite{graves2013generating}. The output
from each of the LSTM layers is ${H = \{\mathbf{h}_t\in \mathbb{R}^k | t = 1, 2, ..., n \}}$.

Next, we concatenate and flatten $H_{forward}$ and $H_{backward}$, obtaining a vector $\mathbf{p} \in \mathbb{R}^{2kn}$. We pass $\mathbf{p}$ through a linear transformation layer and reshape its output to a tensor of size $n \times C \times T$, where $C$ is the number of annotation classes (5 in total, 4 semantic groups and 1 class for negation) and $T$ is the number of possible tags (5 for the IOBES tags). Finally we apply the softmax function along the last dimension of the tensor to approximate the probability for each of the possible tags for each of the annotation class.

\subsection{Word embeddings}


We explored 4 different techniques for learning word embeddings from the text. The embeddings will subsequently be used to initialise the embedding matrix $\mathbf{W}$ that is required by BiLSTM for the NER task. In previous work, the initialisation of $\mathbf{W}$ with pre-trained embeddings has been found to improve the training process \cite{collobert2011natural,mesnil2013investigation}.


\subsubsection*{Random Embeddings} 

Random embeddings were obtained by drawing from a uniform distribution in the ($-0.01$, $0.01$) range. As such, the positions of the words in the vector space do not provide any information regarding patterns of relationships between words.

\subsubsection*{BiLSTM Embeddings} 

These embeddings were obtained after adapting the BiLSTM for a language modelling task. Following a previously described strategy \cite{collobert2008unified}, the input words were randomly replaced, with probability $0.2$, with a word extracted from $V$. We then created a corresponding vector of binary labels to be used as prediction targets: each element of the vector is either $0$ or $1$, where $0$ indicates a word that has been replaced, and $1$ indicates an unchanged word. The model outputs the probability of the labels for each word in the given sentence. After training this language model on the unlabelled part of our corpus, we extracted the word embeddings from $\mathbf{W}$.

\subsubsection*{GloVe Embeddings} 

Word embedding were also obtained using GloVe, an unsupervised method \cite{pennington2014glove}. On word similarity and analogy tasks, it has the potential to outperform competing models such as Skip-gram and CBOW. The GloVe objective function is 
$$
\sum^{|V|}_{i,j=1}f(X_{ij})(w_{i}^{T}\tilde{w}+b_{i}+\tilde{b}_{j}-logX_{ij})^{2}
$$
where $X$ is the word-word co-occurrence matrix, $f$ is a weighting function, $w$ are word embeddings, and $\tilde{w} \in \mathbb{R}^d$ are context word embeddings, with $b$ and $\tilde{b}$ the respective bias terms. The GloVe embeddings $w$ are trained using AdaGrad optimisation algorithm \cite{duchi2011adaptive}, stochastically sampling nonzero elements from $X$.

\subsubsection*{GloVe-Ontology Embeddings} 

Furthermore, we introduced a modified version of GloVe, denoted GloVe-Ontology, with the objective to leverage the RadLex ontology during the word embedding estimation process. The rationale is to impose some constrains on the estimated distance between words using semantic relationships extracted from RadLex; this is an idea somewhat inspired by previous work \cite{yu2014improving}. 

The RadLex data was initially represented as a tree, $\tau$, by considering only the relation \textit{is\_parent\_of} between concepts. We then attempted to string match every word $v$ in $V$ to a concept in $\tau$. Every $v$ matched with a RadLex concept was then assigned the vector that enumerates all ancestors of that concept; otherwise it was associated with a zero vector. We denote the resulting vector by $\phi$. We imposed the constraint that words close to each other in $\tau$ should also be close in the learned embedding space. Accordingly, GloVe's original objective function was modified to incorporate this additional penalty:
$$
\sum^{|V|}_{i,j=1}f(X_{ij})(w_{i}^{T}\tilde{w}+b_{i}+\tilde{b}_{j}-logX_{ij}-\alpha sim(\phi_{i},\phi_{j}))^{2}
$$
In this expression,  $\alpha$ is a parameter controlling the influence of this additional constraint, and $sim$ is taken to be the cosine similarity function. No major changes in the training algorithm were required compared to the original GloVe methodology.
 
\subsection{BiLSTM implementation and training}

The BiLSTM was implemented using two open-source libraries, \textit{Theano} \cite{2016arXiv160502688short} and \textit{Lasagne} \cite{sander_dieleman_2015_27878}. The number of memory cells in each LSTM layer, $k$, was set to $100$. We limited the maximum length of the input sequence to $40$ words and for shorter inputs we used a binary mask at the input and cropped the output predictions accordingly. The loss function was the categorical cross-entropy between the predicted probabilities of the IOBES tags and the true tags. BiLSTM was trained on a GPU for $20$ epochs in batches of $10$ sentences using Stochastic Gradient Descent (SGD) with Nesterov momentum and with the learning rate set to $0.5$. 

The embedding size $d$ was set to $50$. The GloVe, GloVe-Ontology and BiLSTM word embeddings were trained on $743,480$ unlabelled radiology reports. The $\alpha$ paramenter in the Glove-Ontology objective was set to $0.5$.

One aspect of the training was to allow or block the optimisation algorithm from updating the matrix $\mathbf{W}$ in the Embedding Layer of the BiLSTM. In Section \ref{results} we refer to this aspect of training as \textit{fine-tuning}. Previous work \cite{collobert2011natural} has shown that fine-tuning can boost the results of the several supervised tasks in NLP. 

\section{A competing rule-based system}\label{rule_based}



Two clinicians (RB and SW) built a comprehensive dictionary of medical terms. In the dictionary, the key is the name of the term and the corresponding value specifies the semantic group, which was identified using a number of resources. We iterated over all RadLex concepts using the field \textit{Preferred Label} as the dictionary key for the new entry. To obtain the semantic group we traversed up the ontology tree until an ancestor concept was found that had been manually mapped to a semantic group. For example, one of the ancestor concepts of \textit{heart} is \textit{Anatomical entity}, which we had manually mapped to semantic group {\it Body Location}. The same procedure was also performed on the MeSH ontology using the \textit{MeSH Heading} field as a dictionary key. Finally, we added $202$ more terms that were common in day-to-day reporting but were not present in RadLex and MeSH.

The sentences were tokenized and split using the Stanford CoreNLP suite \cite{manning-EtAl:2014:P14-5}, and also converted to lower case and lemmatized using NLTK \cite{bird2009natural}. Next, for each sentence, the algorithm attempted to match the longest possible sequence of words, a target phrase, to an entry in the dictionary of medical terms. When the match was successful, the target phrase was annotated with the corresponding semantic group. When no match was found, the algorithm attempted to look up the target phrase in the English Wikipedia redirects database. In case of a match, the name of the target Wikipedia article was checked against our curated dictionary and the target phrase was annotated with the corresponding semantic group (e.g. \textit{oedema} redirects to \textit{edema}, which is how this concept is named in RadLex).

For all the string matching operations we used SimString \cite{okazaki2010simple}, a fast and efficient approximate string matching tool. We arbitrarily chose the \textit{cosine} similarity measure and a similarity threshold value of 0.85. Using SimString allowed the system to match misspelled  words (e.g. \textit{cardiomegally} to the correct concept \textit{cardiomegaly}).

For negation detection, the system first obtained NegEx predictions for the entities extracted in the NER task. Next, it generated a graph of grammatical relations as defined by the Universal Dependencies \cite{de2014universal} from the Stanford Dependency Parser. It then removed all relations in the graph except \textit{neg}, the negation relation,  and \textit{conj:or}, the \textit{or} disjunction. Given the NegEx output and the reduced dependency graph, the system finally classified an entity as negated if any of the following two conditions were found to be true: (1) any of the words that are part of the entity were in a \textit{neg} relation or in a \textit{conj:or} relation with a another word that was in a \textit{neg} relation; (2) if an entity was classified by NegEx as negated, it was the closest entity to negation trigger and there was no \textit{neg} relations in the sentence. Our hybrid approach is somewhat similar to DEEPEN with the difference that the latter considers all first-order dependency relations between the negation trigger and the target entity. 


\section{Experimental Results} \label{results}

\begin{table}[!t]
\small
\addtolength{\tabcolsep}{-1pt}
\centering
\begin{tabular}{@{}lllll@{}}
\toprule
\textbf{Embeddings} & \textbf{Fine-tuning} & \textbf{P} & \textbf{R} & \textbf{F1} \\ \midrule
Random              & TRUE                 & 0.878      & 0.869      & 0.873      \\
Glove               & TRUE                 & 0.869	     & 0.829	  & 0.849       \\
Glove-ontology      & TRUE                 & 0.875      & 0.860      & 0.867       \\
BiLSTM              & TRUE                 & 0.878      & 0.870      & \textbf{0.874}       \\ \midrule
Random              & FALSE                & 0.829      & 0.727      & 0.775       \\
Glove               & FALSE                & 0.866      & 0.828      & 0.847       \\
Glove-ontology      & FALSE                & 0.850      & 0.839      & 0.844       \\
BiLSTM              & FALSE                & 0.870      & 0.849      & \textbf{0.859}       \\ \midrule
Rule-based     &                      & 0.706      & 0.698      & 0.702       \\ \bottomrule
\end{tabular}
\caption{Comparison of the BiLSTM model and rule-based system. BiLSTM is trained using different word embedding and evaluated using $5$-fold cross-validation. The evaluation considers the overlap span of the semantic group predictions against gold standard annotations.}
\label{overall_results}
\end{table}

\begin{table}[!t]
\small
\centering
\begin{tabular}{@{}lllllll@{}}
\toprule
\textbf{Semantic Group} & \textbf{P} & \textbf{R} & \textbf{F1} \\ \midrule
Body Location           & 0.896              & 0.887           & 0.891       \\
Medical Device          & 0.898              & 0.923           & 0.910       \\
Clinical Finding        & 0.871              & 0.895           & 0.883       \\
Descriptor              & 0.824              & 0.725           & 0.771       \\ \midrule
Total                   & 0.878              & 0.870           & 0.874       \\ \bottomrule
\end{tabular}
\caption{BiLSTM: performance metrics broken down by semantic group for the NER task. All results were obtained using BiLSTM word embeddings.}
\label{results_by_class}
\end{table}

\begin{table}[!t]
\small
\centering
\begin{tabular}{@{}llll@{}}
\toprule
\textbf{Semantic Group} & \textbf{P} & \textbf{R} & \textbf{F1} \\ \midrule
Body Location           & 0.724      & 0.839      & 0.778       \\
Medical Device          & 0.976      & 0.538      & 0.694       \\
Clinical Finding        & 0.862      & 0.551      & 0.672       \\
Descriptor              & 0.467      & 0.780      & 0.584       \\ \midrule
Total                   & 0.706      & 0.698      & 0.702       \\ \bottomrule
\end{tabular}
\caption{Rule-based system: performance metrics broken down by by semantic group for the NER task.}
\label{results_by_class_rule}
\end{table}

\begin{table}[!t]
\small
\centering
\begin{tabular}{lllllll}
\toprule
\textbf{Model}   & \textbf{P} & \textbf{R} & \textbf{F1} \\\midrule
BiLSTM           & 0.903      & 0.912      & 0.908       \\
NegEx            & 0.664      & 0.944      & 0.780       \\
NegEx - Stanford & 0.944      & 0.912      & \textbf{0.928}       \\ \bottomrule
\end{tabular}
\caption{Comparison of BiLSTM, NegEx and NegEx-Stanford for negation detection. All algorithms predicted whether a given medical entity was negated or affirmed.}
\label{negation_results}
\end{table}

\begin{table}[!t]
\centering
\small
\addtolength{\tabcolsep}{-6pt}
\begin{tabular}{ccccc}
\toprule
\textbf{nodule} & \textbf{pacemaker} & \textbf{small} & \textbf{remains} & \textbf{fracture} \\ \hline
bulla           & ppm                & tiny           & remain           & fractures         \\
nodules         & icd                & minor          & appears          & deformity         \\
opacity         & wires              & mild           & is               & body              \\
opacities       & drains             & dense          & are              & scoliosis         \\
opacification   & leads              & extensive      & were             & abnormality       \\ \bottomrule
\end{tabular}
\caption{For each one of the five words in boldface, five nearest neighbours found in the embedding space learnt by BiLSTM.}
\label{embeddings_example}
\end{table}



We evaluated the BiLSTM model on the medical NER task by measuring the overlap between the predicted semantic groups and the ground truth labels. The evaluation was performed at the granularity of a single word and using $5$-fold cross-validation. The BiLSTM model was always trained on $80\%$ of the annotated corpus and tested on the remaining $20\%$.

Table \ref{overall_results} compares the performance of various BiLSTM variants that were obtained with and without fine-tuning of the word embeddings to the performance of our baseline rule-based system. Without fine-tuning, the BiLSTM NER model, that was initialised with the embeddings trained in an unsupervised manner using the BiLSTM language model, achieves the best F1-score ($0.859$), and outperforms the next best variant by $0.012$. With fine-tuning, the same BiLSTM variant improves the F1-score by a further $0.015$ and outperforms the baseline rule-based system by an F1-score of $0.172$. Table \ref{results_by_class} shows its performance measure for each of the semantic groups.

The evaluation of negation detection was measured on complete entities. If any of the words within an entity were tagged with a I, B, E or S, that entity was considered to be negated. As shown in Table \ref{negation_results}, the BiLSTM (BiLSTM language model embeddings, fine-tuning allowed) achieved an F1-score of $0.902$, which outperformed NegEx by $0.128$. However, the best F1-score of $0.928$ is achieved using the NegEx-Stanford system.

\section{Discussion}

In Table \ref{results_by_class}, we show the predictive performance of the best BiLSTM NER model for each of the semantic groups. \textit{Body Location}, \textit{Medical Device} and \textit{Clinical Finding} show a balanced precision and recall, and similar F1-scores. \textit{Descriptor} has a lower F1-score which is caused by a low recall that may be the results of the larger variability in the words used for this semantic group. Table \ref{results_by_class_rule} shows the corresponding results for the rule-based NER system. \textit{Medical Device} and \textit{Clinical Finding} show a typical performance for a dictionary-based NER system with a high precision and a low recall. \textit{Body Location} has relatively high precision and recall values which suggests that this semantic group is well covered by our dictionary of medical terms. In contrast, \textit{Descriptor} shows a very low precision which is the result of a high number of false positives. The false positives are caused by many \textit{Descriptor} entries in our dictionary of medical terms that had been automatically extracted from RadLex and MeSH but which do not correspond to the definition of a \textit{Descriptor} used by the clinicians who produced the labelled data. 

As a qualitative assessment, Table \ref{embeddings_example} shows the 5 nearest neighbours obtained from BiLSTM language model embeddings of some frequent words used by Radiologists. We note that there is an clear semantic similarity between the nearest neighbour words. Additionally, the embeddings encode syntactic information as the nearest neighbour words are parts of speech of the same type as the target word. We also summed the vectors for \textit{heart} and \textit{enlarged}, which yielded vec(\textit{cardiomegaly}) as the nearest vector. Similarly, the closest vector to vec(\textit{heart}) + vec(\textit{not}) + vec(\textit{enlarged}) is vec(\textit{normal}). These examples suggest that word embeddings may encode information about the compositionality of words as discussed by Mikolov \shortcite{mikolov2013distributed}. 


Table \ref{overall_results} shows that, without fine-tuning, the Embedding Layer weights can affect the performance of the NER task. When fine-tuning is allowed there is only a marginal advantage in using pre-trained embeddings, as the BiLSTM performs equally well when initialised with random embeddings. Therefore, despite a positive qualitative assessment, the pre-trained word embeddings seem to offer only a small advantage when used for the proposed NER task as  BiLSTM is able to learn well using the annotated data during the supervised learning phase.

\section{Conclusions} \label{conclusions}

In this paper we have shown that a recurrent neural network architecture, BiLSTM, can learn to detect clinical findings and negations using only a relatively small amount of manually labelled radiological reports. Using a manually curated medical corpus, we have provided initial evidence that BiLSTM outperforms a dictionary-based system on the NER task. For the detection of negations, on our dataset BiLSTM approaches the performance of a negation detection system that was build using the popular NegEx algorithm and uses grammatical relations obtained from the Stanford Dependency Parser and hand-crafted rules. We believe that increasing the size of the annotated training dataset can result in much improved performance on this task, and plan to purse this investigation in future work. 

We have also investigated potential performance gains that can be achieved by using pre-trained word embeddings, i.e. BiLSTM, GloVe and GloVe-Ontology embeddings, in the context of BiLSTM-based modelling for the NER task. Our initial experimental results suggest that there is marginal benefit in using BiLSTM-learned embeddings while pre-training using GloVe and GloVe-Ontology embeddings did not offer any significant improvements over a random initialisation. 

\bibliography{mybibfile}
\bibliographystyle{emnlp2016}

\end{document}